%% file: main.tex
\documentclass[final]{siamltex}
\usepackage{algorithm,algpseudocode}
\usepackage{latexsym, graphicx, epsfig, amsmath, amsfonts,amssymb,bm}
\usepackage{caption, subcaption}
\usepackage{epstopdf}
\usepackage{booktabs}
\usepackage{array}
\usepackage{color}
\usepackage{lineno}
\usepackage{url}
\usepackage[version=4]{mhchem}
\usepackage{fixmath}
\usepackage{diagbox}

\def\Dt{\Delta t}

\allowdisplaybreaks

\def\be{\begin{equation}}
\def\ee{\end{equation}}

\def\x{\mathbf{x}}

\def\f{\mathbf{f}}

\def\z{\mathbf{z}}

\def\N{\mathbf{N}}

\def\PPh{\boldsymbol{\Phi}}
\def\PPs{\boldsymbol{\Psi}}

\def\R{{\mathbb R}}

\def\I{\mathbf I}

\def\w{\mathbf{w}}

\def\0{\mathbf{0}}

\title{Deep Learning of Chaotic Systems from Partially-Observed Data}

\author{Victor Churchill\footnotemark[1]\thanks{Department of Mathematics,
		The Ohio State University, Columbus, OH 43210, USA. Emails:
		{\tt churchill.77@osu.edu, xiu.16@osu.edu} Funding: This 
		work was partially supported by AFOSR FA9550-22-1-0011.}\and Dongbin Xiu\footnotemark[1]
				}

\begin{document}
\maketitle
\begin{abstract}
Recently, a general data driven numerical framework has been developed for learning and modeling of unknown dynamical systems using fully- or partially-observed data. The method utilizes deep neural networks (DNNs) to construct a model for the flow map of the unknown system. Once an accurate DNN approximation of the flow map is constructed, it can be recursively executed to serve as an effective predictive model of the unknown system. In this paper, we apply this framework to chaotic systems, in particular the well-known Lorenz 63 and 96 systems, and critically examine the predictive performance of the approach. A distinct feature of chaotic systems is that even the smallest perturbations will lead to large (albeit bounded) deviations in the solution trajectories. This makes long-term predictions of the method, or any data driven methods, questionable, as the local model accuracy will eventually degrade and lead to large pointwise errors. Here we employ several other qualitative and quantitative measures to determine whether the chaotic dynamics have been learned. These include phase plots, histograms, autocorrelation, correlation dimension, approximate entropy, and Lyapunov exponent. Using these measures, we demonstrate that the flow map based DNN learning method is capable of accurately modeling chaotic systems, even when only a subset of the state variables are available to the DNNs. For example, for the Lorenz 96 system with 40 state variables, when data of only 3 variables are available, the method is able to learn an effective DNN model for the 3 variables and produce accurately the chaotic behavior of the system.
\end{abstract}
\begin{keywords}
Deep neural networks, chaotic behavior, flow map
\end{keywords}

\input Introduction
\input Setup
\input ComputationalStudies
\input Results
\input Conclusion

\bibliographystyle{siamplain}
\bibliography{neural,LearningEqs,ensemble,lorenz}

\end{document}

%% file: Introduction.tex
\section{Introduction} \label{sec:intro}

Due to recent advances in machine learning software and computing hardware combined with the availability of vast amounts of data, data-driven learning of unknown dynamical systems has been a very active research area in the past few years. One way to approach this problem is governing equation discovery, where a map is constructed from state variables to their time derivatives. Among other techniques, this can be achieved via sparse approximation, where under certain circumstances exact equation recovery is possible. See, for example, \cite{brunton2016discovering} and its many extensions in recovering both ODEs in \cite{brunton2016discovering,kang2019ident, schaeffer2017sparse,schaeffer2017extracting, tran2017exact} and PDEs in \cite{rudy2017data, schaeffer2017learning}. Deep neural networks (DNNs) have also been used to construct this mapping. See, for example, ODE modeling in \cite{lu2021deepxde,qin2018data,raissi2018multistep,rudy2018deep}, and PDE modeling in \cite{long2018pde,long2017pde,lu2021learning,raissi2017physics1,raissi2017physics2,raissi2018deep,sun2019neupde}. 

Another approach for learning unknown systems, our focus here, is flow map or evolution discovery, where a map is constructed between two system states separated by a short time to approximate the flow map of the system, \cite{qin2018data}. Unlike governing equation discovery, approximating the flow map does not directly yield the specific terms in the underlying equations. Rather, if an accurate flow map is discovered, then an accurate predictive model can be defined for the evolution of the unknown system such that a new initial condition can be marched forward in time. This approach relies on a DNN, in particular residual network (ResNet) \cite{he2016deep}, to approximate the flow map. Since its introduction in \cite{qin2018data} to model autonomous systems, a general framework has been developed for the flow map approximation of unknown systems from their trajectory data that extends to non-autonomous systems \cite{QinChenJakemanXiu_SISC}, parametric dynamical systems
\cite{QinChenJakemanXiu_IJUQ}, partially observed dynamical systems
\cite{FuChangXiu_JMLMC20}, as well as partial differential equations
\cite{chen2022deep,WuXiu_modalPDE}. Of particular interest in this paper is \cite{FuChangXiu_JMLMC20}, where a finite memory of the state variable time history is used to learn  reduced systems where only some of the state variables are observed per the Mori-Zwanzig formulation. 

The focus of this paper is to extend this flow map deep learning framework to {\em chaotic systems} and examine its performance for both fully and partially observed chaotic systems. Chaotic systems exhibit ultra sensitivity to perturbations of the system parameters and initial conditions. Hence, they represent a highly challenging case for any learning and modeling methods, particularly for long-term system behavior. Although well known chaotic systems, e.g., the Lorenz 63 system, have been adopted in the literature, they were mostly used as demonstrative examples of the proposed methods in the papers by using visual examination of phase plots. The nearly impossible task of matching the long term evolution of the true chaotic system is rarely addressed in the existing literature. With an exclusive focus on chaotic systems, the purpose of this paper is to systematically examine the long term predictive accuracy using a set of measures beyond phase plots. These include bounded pointwise error, histograms, autocorrelation functions, correlation dimension, approximate entropy, and Lyapunov exponent. Using these measures, we further establish that the flow map based deep learning method is capable of learning and modeling chaotic systems and producing accurate long term system predictions, for both fully- and partially-observed systems.

\subsection{Literature Review}
The problem of learning fully- and partially-observed chaotic systems, especially the famous Lorenz 63 system, has been a popular topic particularly in the age of machine learning. Hence, we limit the scope of this review to papers particularly relevant to the subject of learning chaotic dynamics.

The recent paper \cite{bhat2022recurrent} is most similar to this work in that the authors seek to learn partially-observed chaotic dynamics using memory. This paper considers observing just one variable while we consider many combinations of partially-observed variables, and focuses on optimizing the network parameters (e.g. number of neurons and memory length) of a particular recursive structure based on root mean squared error (RMSE), while we consider additional measures of chaotic dynamics and a general ResNet. There are also several other relevant papers including \cite{trischler2016synthesis}, which uses a recurrent neural network to predict chaotic systems including Lorenz 63. In \cite{wulkow2021data}, the authors discuss memory length and Takens' theorem \cite{takens1981detecting} in the context of learning Lorenz 96. In addition, \cite{scher2019generalization} considers approximating chaotic dynamical systems using limited data and external forcing. Also, \cite{vlachas2018data,pawar2021nonintrusive,chattopadhyay2020data,dubois2020data} use long short-term memory (LSTM) recurrent networks to study fully- and partially-observed Lorenz systems.

There are also several earlier papers, including \cite{zimmermann2000modeling} which proposes modeling dynamical systems via recurrent neural networks, \cite{kim1999nonlinear} which deals with estimating time delays (memory length) in Lorenz and other systems, \cite{bakker2000learning} that learns chaotic dynamics of reduced systems via neural networks, \cite{miyoshi1995learning} which learns chaotic dynamics with neural networks, and \cite{han2004modeling,han2004prediction}, which predict the chaotic Rossler system using a simple RNN. These papers are certainly relevant, but lack the extensive examples and systematic verification experiments that are now more easily achieved with today's computing systems.

It is also worth mentioning a large body of work that uses a different approach (governing equation discovery) to learning chaotic systems. In \cite{brunton2016discovering}, the authors approximate the particular terms in chaotic systems (the right hand sides) through sparse optimization or network learning. Other work in this area includes \cite{brunton2017chaos,lusch2018deep} which focus on learning delayed embeddings, \cite{raissi2018multistep} which uses physics-informed neural networks, \cite{pan2018data}, which explicitly learns the closure of the partially observed Lorenz 63 system via a NN, \cite{champion2019data} which combines learning a coordinate transform from the delayed embedding coordinates with learning the dynamic coordinates of chaotic systems, \cite{rudy2019deep} which learns the Lorenz system from noisy data, and \cite{bakarji2022discovering} which learns the governing equations for Lorenz (or a Lorenz-like surrogate) from just one observation variable.

\subsection{Contributions}


The chief contribution of this paper is a systematic and rigorous examination of learning the flow maps of fully- and partially-observed chaotic dynamical systems using an approachable and mathematically grounded DNN framework. In most papers dealing with this topic, pointwise error or a visual comparison of phase plots are used to assess the accuracy of network prediction. However, it is well-known that chaotic systems are extremely sensitive to perturbation, and therefore in the long run the model predictions will dramatically stray from the truth. This makes assessing the efficacy of the learned system, particularly for long-term, a challenging problem in and of itself. Hence, part of our contribution is the proposal of a more robust approach to the assessment of learning chaotic behavior not explored in the existing literature that includes standard techniques of error analysis such as pointwise error and phase plots as well as comparison of a variety of other measures that demonstrate accurate behavior including matching of histograms and autocorrelation, and statistics that quantify chaos such as correlation dimension, approximate entropy, and Lyapunov exponent. Finally, we contribute several ambitious numerical examples not explored in the literature. As a benchmark, we consider learning the  well-known Lorenz systems with different combinations of observed variables. Of particular note is a 40-dimensional Lorenz 96 system with only 3 variables observed, where we demonstrate that the DNN method can learn the chaotic behavior in general despite training data being collected from a single long trajectory consisting of only 3 variables (out of 40).


%% file: Setup.tex
\section{Flow Map Modeling of Unknown Chaotic Dynamical Systems} \label{sec:setup}

We are interested in constructing effective models for the evolution
laws behind chaotic dynamical data. We follow the framework in \cite{qin2018data,FuChangXiu_JMLMC20} for learning fully- and partially-observed dynamical systems via residual DNNs.
%
The following review closely follows that of \cite{churchill2022robust}. Throughout this paper our discussion will be on dynamical systems observed over discrete time
instances with a constant time step $\Dt$,
\be \label{tline}
t_0<t_1<\cdots, \qquad
t_{n+1} - t_n = \Dt, \quad \forall n.
\ee
Generality is not lost with the constant time step assumption, as
variable time step can be treated as a separate entry to the DNN
structure \cite{QinChenJakemanXiu_SISC}.
We will use a subscript to denote the time variable of
a function, e.g., $\x_n = \x(t_n)$.


\subsection{ResNet Modeling of Fully-Observed Systems}

Consider an unknown autonomous system,
\be\label{eq:ODE}
\frac{d\x}{dt} = \f(\x), \qquad \x\in\R^d,
\ee
where $\f:\R^d\to \R^d$ is not known. Because it is autonomous, its flow map depends only
on the time difference as opposed to the actual time, i.e.,
$ \x_n = \PPh_{t_n-t_s}(\x_s)$. Thus, the solution having been marched forward one time step
satisfies
\be
\x_{n+1} = \PPh_{\Dt}(\x_n) = \x_n + \PPs_{\Dt}(\x_n),
\ee
where $\PPs_{\Dt} = \PPh_{\Dt} - \mathbf{I}$, with $\mathbf{I}$ as the
identity operator.

When data for all of the state variables $\x$ over the time stencil
\eqref{tline} are available, they can be grouped into sequences
$$
\{\x^{(m)}(0), \x^{(m)}(\Dt),\ldots,\x^{(m)}(K\Dt)\},\quad m=1,\ldots,M,
$$
where $M$ is the total number of such data sequences and $K+1$ is the
length of each sequence (which is assumed to be a constant for
notational convenience). This serves as the training data set.
Inspired by basic numerical schemes for solving ODEs, one can model the unknown evolution operator using a residual network (ResNet) (\cite{he2016deep}) in the
form of
\be
\mathbf{y}^{out} = \left[\mathbf{I}+\mathbf{N} \right](\mathbf{y}^{in}),
\ee
where $\N:\R^d\to\R^d$ stands for the mapping operator of a standard
feedforward fully connected neural network.
The network is then trained by using the training data set 
and minimizing the recurrent mean squared loss function
\be
\frac1M\sum_{m=1}^M\sum_{k=1}^K \left\| \x^{(m)}(k\Dt) - [\I+\N]^k(\x^{(m)}(0))\right\|^2,
\ee
where $[\mathbf{I}+\mathbf{N}]^k$ indicates composition of the network function $k$ times. Recurrent loss is used to increase the stability of the network approximation over long term prediction. The trained network thus accomplishes
$$
\x^{(m)}(k\Dt) \approx [\I+\N]^k(\x^{(m)}(0)), \qquad \forall m=1,\ldots,M, ~k=1,\ldots,K.
$$
After the network is trained to a satisfactory accuracy level, it can then be used as a predictive model
\be \label{ResNet}
\x_{n+1} =  \x_n + \N(\x_n), \qquad n=0,1,\dots,
\ee
starting from any initial condition $\mathbf{x}(t_0)$. This framework was
proposed in  \cite{qin2018data} and was extended to parametric systems
and time-dependent (non-autonomous) systems (\cite{QinChenJakemanXiu_IJUQ,QinChenJakemanXiu_SISC}).

\subsection{Memory-based ResNet Modeling of Partially-Observed Systems}\label{subsec:memory}

A notable extension of the flow-map modeling in \cite{qin2018data} is
to partially-observed systems where some state variables are not observed at all. Let $\x = (\z; \w)$, where
$\z\in\R^m$ and $\w\in\R^{d-m}$. Let $\z$ be the observables and $\w$
be the missing variables. That is, no information or data of $\w$ are
available. When data are available only for $\z$, it is possible
to derive a system of equations for $\z$ only via the Mori-Zwanzig (MZ)
formulation  (\cite{mori1965, zwanzig1973}). However, the MZ
formulation asserts that the reduced system for $\z$ requires a
memory integral, whose kernel function, along with other terms in the formula, is unknown. By making a mild assumption that the memory is of
finite (problem-dependent) length, memory-based DNN
structures were investigated in \cite{WangRH_2020, FuChangXiu_JMLMC20}.
While \cite{WangRH_2020} utilized LSTM (long short-term memory) networks,
\cite{FuChangXiu_JMLMC20} proposed a relatively simple DNN structure, in
direct correspondence to the Mori-Zwanzig formulation,
that takes the following mathematical form,
\be\label{eq:memory}
      \z_{n+1} = \z_n +
      \N(\z_n,\z_{n-1}, \dots,\z_{n-n_M}),  \qquad n\geq n_M,
      \ee
      where $n_M\geq 0$ is the number of memory terms in the model. In
      this case, the DNN operator is $\N:\R^{m\times (n_M+1)}\to
      \R^m$, which corresponds to a ResNet with additional time history inputs. The special case of $n_M=0$ corresponds to the standard
      ResNet model \eqref{ResNet} for modeling fully-observed systems
      without missing variables (thus no need for memory).
      
In the case of $m=1$, Takens' theorem \cite{takens1981detecting} proves (non-constructively) the existence of a map from $F:\mathbb{R}^{n_M+1}\rightarrow \mathbb{R}$ such that $z_{n+1} = F(z_{n},\ldots,z_{n-n_M})$ provided $n_M\ge2d$. This serves as inspiration that, given enough data, a universal approximator can find this map. However, in this paper we consider many values for $m$ corresponding to different combinations of observed variables.

%% file: ComputationalStudies.tex
\section{Computational Framework}

In the main task of this paper, we apply the flow map learning methods described in the previous section to chaotic systems. First, we review the setting including the DNN structure used as well as the specifics of the data generation and model training. Next, we discuss the qualitative and quantitative metrics to evaluate the flow map learning. We then present extensive numerical results for learning fully- and partially-observed Lorenz systems.

\subsection{DNN Structure}

\begin{figure}[htbp]
	\begin{center}
		\includegraphics[width=\textwidth]{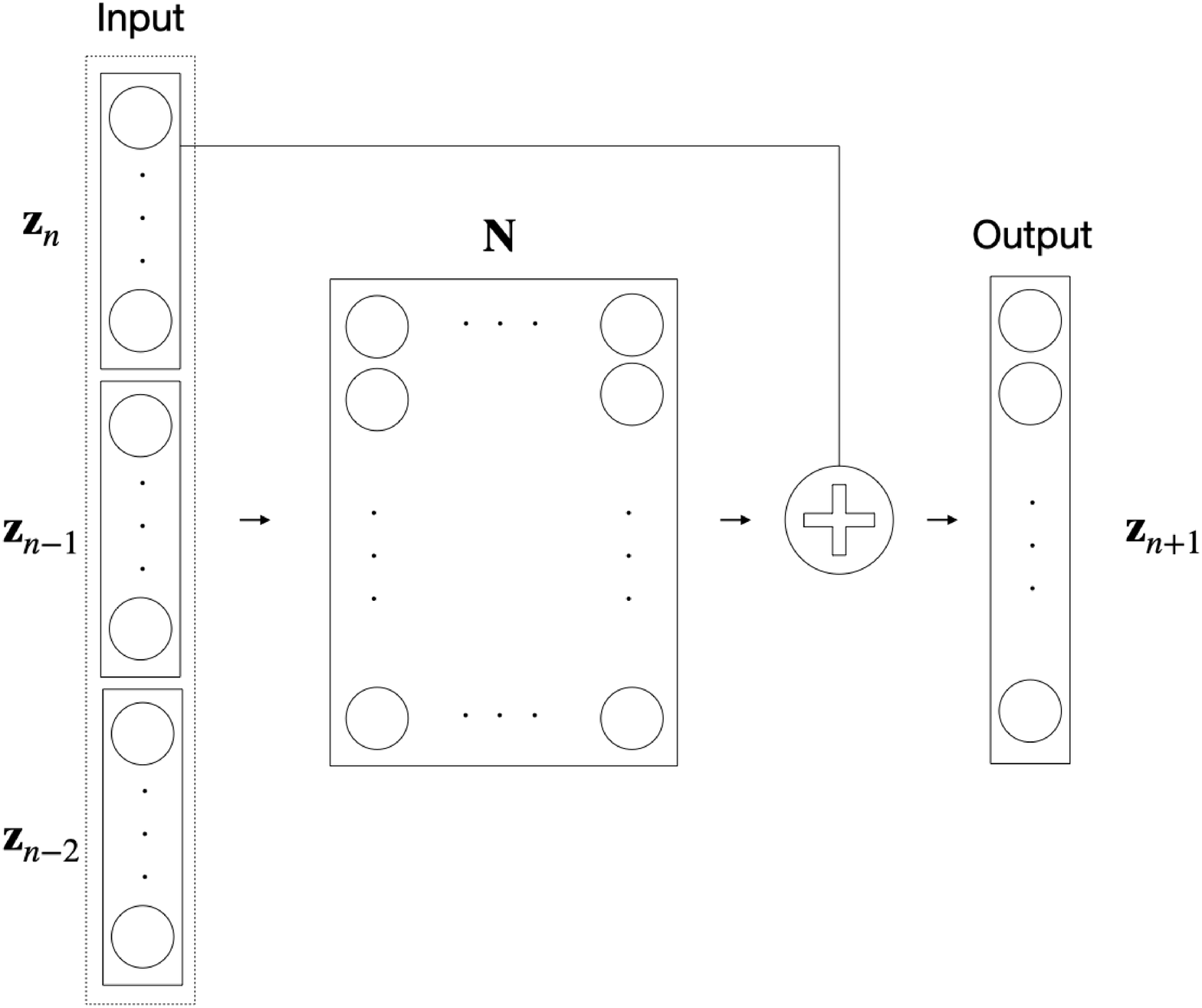}
		\caption{Memory-Based DNN Structure.}
		\label{fig:structure}
	\end{center}
\end{figure}

The structure of the DNNs used to achieve chaotic flow map learning is modeled by \eqref{eq:memory} (\cite{FuChangXiu_JMLMC20}). The network function $\mathbf{N}:\mathbb{R}^{m\times(n_M+1)}\rightarrow\mathbb{R}^m$ maps the input time history of the observed variables to the output future time through a series of fully-connected (also known as dense) layers with ReLU activation. An illustration of this structure with memory length $n_M=2$ is shown in Figure \ref{fig:structure}. Note that this structure is general in that while it is built for partially-observed systems that require memory based on the Mori-Zwanzig formulation, setting the memory length $n_M=0$ returns a standard ResNet that is appropriate for learning fully-observed systems. Our extensive numerical experimentation and previous work with this framework indicate that particularly wide or deep networks are not typically necessary for learning with this structure, and hence most examples in this paper use $3$ hidden layers with $20$ neurons in each layer. The choice of memory length $n_M$ is in general problem-dependent and hinges on a number of factors including the time step and the relationship between the observed and missing variables, and is explored in detail in \cite{FuChangXiu_JMLMC20}. In the examples in this paper, typically $n_M=10$ time steps (equivalent to $0.1$ seconds in time as discussed below). 

\subsection{Data Generation and Model Training}
For benchmarking purposes, in all examples the true chaotic systems we seek to approximate are known. However, these true models serve only two purposes: (1) to generate synthetic data with which to train the DNN flow map approximations; and (2) to generate reference solutions for comparison with DNN predictions in testing. Therefore, the knowledge of the true system does not in any way facilitate the DNN model approximation.

Data generation for both of these tasks is achieved by solving the true systems using a high-order numerical solver, and observing this reference solution at discrete time steps with $\Dt = 0.01$ seconds. To generate the training data, a single initial condition $\x(0)$ generates a long trajectory from $t=0$ to $t=10,000$ seconds ($1,000,000$ time steps). From this long trajectory, $M$ sequences of length $n_M+K+1$ are collected uniformly at random to form the training data set
\be\label{eq:dataset}
\{\x^{(m)}_{n-n_M},\ldots, \x^{(m)}_{n-1},\x^{(m)}_n,\x^{(m)}_{n+1},\ldots,\x^{(m)}_{n+K}\},\quad m=1,\ldots,M,
\ee
where $n_M$ is the memory length and $K$ is the recurrent loss parameter. In our examples, typically $M=10,000$ and $K=10$ time steps (equivalent to $0.1$ seconds). For systems requiring memory, data is generated from the full true system for all state variables $\mathbf{x}$, with only the data for the observed state variables $\mathbf{z}$ being kept and data for the missing variables $\mathbf{w}$ being discarded.

Once the training dataset has been generated, the learning task is achieved by training the DNN model \eqref{eq:memory} with the data set \eqref{eq:dataset}. In particular, the network hyperparameters (weights and biases) are trained by minimizing the recurrent loss function
\be
\frac1M\sum_{m=1}^M\sum_{k=1}^K \left\| \x^{(m)}_{n+k} - [\I_n+\N]^k(\x^{(m)}_n,\ldots,\x^{(m)}_{n-n_M})\right\|^2,
\ee
where $\I_n(\x_n,\ldots,\x_{n-n_M})=\x_n$, using the stochastic optimization method Adam \cite{kingma2014adam}. In the examples below we typically train for $10,000$ epochs with batch size $50$ and a constant learning rate of $10^{-3}$ in Tensorflow \cite{tensorflow2015}.

\subsection{DNN Prediction and Validation}

After satisfactory network training, we obtain a predictive model \eqref{eq:memory} for the unknown system which can be marched forward in time from any new initial condition. To validate the network prediction, testing data is generated in the same manner as training data was above. In particular, a \emph{new} initial condition generates a reference solution from $t=0$ to $t=100$ seconds ($10,000$ time steps) using the true governing equations. For prediction, the DNN model is marched forward starting with the first $n_M+1$ time steps of the test trajectory until $t=100$ and is compared against the reference. Note that we march forward significantly longer than $K\Dt$ (the length of each training sequence which is typically $t=0.1$) to examine the long-term system behavior.

For chaotic systems, it is practically impossible to approximate the flow map with a model that achieves low pointwise error in the long term due to the fact that even machine epsilon changes in the initial condition or other system parameters will drastically change the evolution of the system. However, these changes will not alter the physics of the system, i.e. the nature of the behavior of the system. Hence, in this study, we recognize the challenge of low long-term pointwise error and focus on learning the physics of the system to match the chaotic behavior. In particular, we look at a myriad of qualitative and quantitative metrics in order to demonstrate a strong match of physics including: pointwise error, phase plot, histogram, autocorrelation function, correlation dimension, approximate entropy, and Lyapunov dimension. These tools have been used to classify chaotic behavior in Lorenz and other chaotic systems, e.g. in \cite{bakker2000learning,chattopadhyay2020data,kim1999nonlinear,rudy2019deep}. We briefly review each of the evaluation tools below. All metrics are computed using MATLAB \cite{MATLAB}.

\begin{itemize}
\item Pointwise error: In the following examples, we look at the reference test trajectories versus the trajectories predicted by the network model. However, the predictions will quickly deviate from the reference trajectories since the dynamics we learn are in fact an approximation and at best some small perturbation away from the true dynamics.\footnote{For that matter, even the reference trajectories themselves are just an approximation of the unknown true trajectories as they are generated with a high-order numerical solver rather than analytically solving the system.} For a chaotic system, this small perturbation causes drastic changes, which stay bounded, in the system later in time. Hence, we can also look at the log absolute error between the trajectories, and check that this quantity remains bounded. Bounded pointwise error demonstrates stability of the predictive model.

\item Phase plot: We also qualitatively compare the phase plots of reference and predicted test trajectories when multiple state variables are observed. This can serve as a qualitative measure of the behavior of the system, e.g. if both systems exhibit two attractors centered at the same particular points.

\item Histogram: We compare the approximate densities of reference and predicted values for the trajectories as well, where a match in the distribution indicates similar behavior in the long term. The histograms are computed over all time steps starting with the initial condition.

\item The \textbf{autocorrelation function} (\cite{box2015time}) measures the correlation between the time series $x_t$ and its lagged counterpart $x_{t+k}$, where $k=0,\ldots,K$ and $x_t$ is a stochastic process. The autocorrelation for lag $k$ is defined as $r_k = c_k/c_0$ where
\begin{align}
c_k = \frac1T \sum_{t=1}^{T-k} (x_t-\bar{x})(x_{t+k}-\bar{x})
\end{align}
and $c_0$ is the sample variance of the time series, with $\bar{x}$ the mean of the time series and $T$ the length of the time series.
In our implementation, the sample autocorrelation is computed by MATLAB's Econometrics Toolbox \cite{matlabecon} using the default settings. See this link\footnote{https://www.mathworks.com/help/econ/autocorr.html} for further computational details.

\item The \textbf{correlation dimension}  (\cite{theiler1987efficient}) is a measure of chaotic signal complexity in multidimensional phase space. Specifically, it is a measure of the dimensionality of the space occupied by a set of random points, whereby a higher correlation dimension represents a higher level of chaotic complexity in the system. It is computed by first generating a delayed reconstruction $Y_{1:N}$ with embedding dimension $m$ (in our implementation $m$ is set to be the dimension of the full system regardless of whether the system is partially- or fully-observed) and lag $\tau$ of reference or predicted trajectories assembled in a matrix $X$. The number of with-in range points, at point $i$, is calculated by
\begin{align}\label{eq:withinrange}
N_i(R) = \sum_{i=1,i\neq k}^N \mathbf{1}(\|Y_i-Y_k\|_\infty<R)
\end{align}
where $\mathbf{1}$ is the indicator function, $R$
is the radius of similarity, and $N$ is the number of points used to compute $R$.
The correlation dimension is the slope of $C(R)$ vs. $R$, where the correlation integral is defined as
\begin{align}
C(R) = \frac{2}{N(N-1)} \sum_{i=1}^N N_i(R).
\end{align}
In our implementation, the correlation dimension is computed by MATLAB's Predictive Maintenance Toolbox \cite{matlabpredmaint} using default settings (e.g. for $N$ and $R$). See this link\footnote{https://www.mathworks.com/help/predmaint/ref/correlationdimension.html} for further computational details.

\item The \textbf{approximate entropy} (\cite{pincus1991approximate}) is a measure used to quantify the amount of regularity and unpredictability of fluctuations over a nonlinear time series. It is computed as
$\Phi_m-\Phi_{m+1}$, where
\begin{align}
\Phi_m = (N-m+1)^{-1}\sum_{i=1}^{N-m+1}\log(N_i(R)),
\end{align}
where $m$, $N$, $R$, and $N_i(R)$, are defined as above when discussing correlation dimension using the same delayed reconstruction. In our implementation, the approximate entropy is computed by MATLAB's Predictive Maintenance Toolbox \cite{matlabpredmaint} using default settings. See this link\footnote{https://www.mathworks.com/help/predmaint/ref/approximateentropy.html}, which contains a working example of the Lorenz 63 system, for further computational details.

\item The \textbf{Lyapunov exponent} (\cite{rosenstein1993practical}) characterizes the rate of separation of infinitesimally close trajectories in phase space to distinguish different attractors, which can be useful in quantifying the level of chaos in a system. A positive Lyapunov exponent indicates divergence and chaos, with the magnitude indicating the rate of divergence. For some point $i$, the Lyapunov exponent is computed using the same delayed reconstruction $Y_{1:N}$ as the correlation dimension and approximate entropy as
\be
\lambda(i) = \frac{1}{(K_{max}-K_{min}+1)dt}\sum_{K=K_{min}}^{K_{max}} \frac1K \ln \frac{\|Y_{i+K}-Y_{i^*+K}\|}{\|Y_{i}-Y_{i^*}},
\ee
where $K_{min}$ and $K_{max}$ represent the expansion range, $dt$ is the sample time (equal to $\Delta t=0.01$ in our case), and $i^*$ is the nearest neighbor point to $i$ satisfying a minimum separation. A single scalar for the Lyapunov exponent is then computed as the slope of a linear fit to the $\lambda(i)$ values. In our implementation, the Lyapunov exponent is computed by MATLAB's Predictive Maintenance Toolbox \cite{matlabpredmaint} using sampling frequency of $100$ (corresponding to $\Dt =0.01$ and otherwise default settings.
See this link\footnote{https://www.mathworks.com/help/predmaint/ref/lyapunovexponent.html}, which also contains a working example of the Lorenz 63 system, for computational details.
\end{itemize}


We note that while the computational specifics of each of these evaluation tools are in fact tunable and indeed changing them would yield different values, the comparisons that follow are not confined to accuracy for these particular default implementation choices and we present them simply as an example of one configuration.

%% file: Results.tex
\section{Computational Results}

In this section we provide numerical examples of DNN modeling of
chaotic systems. We focus on two well-known systems: the 3-dimensional
Lorenz 63 system and the 40-dimensional Lorenz 96 system. In each system,
we start with the learning of fully-observed variables to demonstrate
the effectiveness of ResNet learning. We then focus on
partially-observed cases. For the 3-dimensional Lorenz -63 system, we
examine the DNN learning using training data of (different
combinations of) only 2 variables, as well as of only 1 variables. For
the 40-dimensional Lorenz-96 system, we examine the DNN learning when
only 3 variables are observed in the training data.

\subsection{Low-dimensional system: Lorenz 63}

The Lorenz 63 system is a nonlinear deterministic chaotic 3-dimensional system 
\begin{align}
\begin{split}
\frac{dx}{dt} &= \sigma(y-x),\\
\frac{dy}{dt} &= x(\rho-z)-y,\\
\frac{dz}{dt} &= xy-\beta z,
\end{split}
\end{align}
where $\sigma$, $\rho$, and $\beta$ are parameters. It was proposed in
\cite{lorenz1963deterministic} as a simplified model for atmospheric
convection. When $\sigma=10$, $\rho=28$, and $\beta=8/3$, the system
exhibits chaotic behavior and is a widely studied case.

\subsubsection{Example 1: Full 3-dimensional system}

In this example all three state variables $x$, $y$, and $z$ are
observed and stored in the training data set. In particular, a
trajectory is generated by a high-order numerical solver and observed
at every $\Delta t=0.01$ starting from initial condition
$(x_0,y_0,z_0) = (1,1,1)$ until $T=10,000$ seconds. From this long
trajectory, $10,000$ data sequences of length $0.1$ seconds ($11$ time
steps) are taken as training data, which allows $10$ steps to be used
for recurrent loss. A standard ResNet with $3$ hidden layers with $20$
neurons each is used, and the mean squared recurrent loss function is
minimized using Adam with a constant learning rate of $10^{-3}$ for
$10,000$ epochs. 

\begin{figure}[t]
	\begin{center}
		\includegraphics[width=\textwidth]{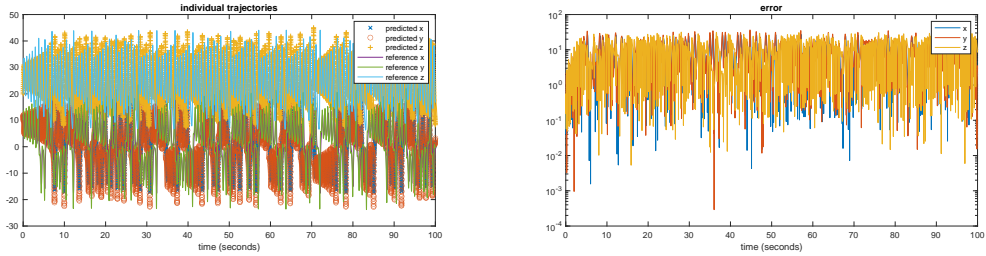}
		\caption{Ex. 1: Lorenz 63 Full System -- Individual trajectory comparison and pointwise error.}
		\label{fig:lorenz63full_error}
	\end{center}
\end{figure}

\begin{figure}[t]
	\begin{center}
		\includegraphics[width=\textwidth]{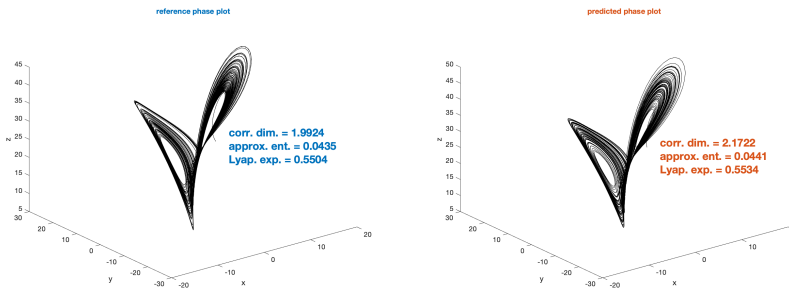}
		\caption{Ex. 1: Lorenz 63 Full System -- Phase
                  plots. (Left: reference; Right: DNN prediction.)}
		\label{fig:lorenz63full_phase}
	\end{center}
\end{figure}

\begin{figure}[t]
	\begin{center}
		\includegraphics[width=\textwidth]{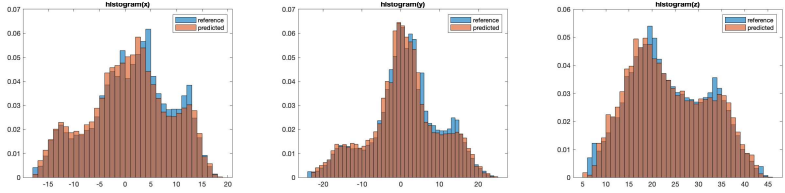}
		\caption{Ex. 1: Lorenz 63 Full System -- Histogram comparison.}
		\label{fig:lorenz63full_hist}
	\end{center}
\end{figure}

\begin{figure}[t]
	\begin{center}
		\includegraphics[width=\textwidth]{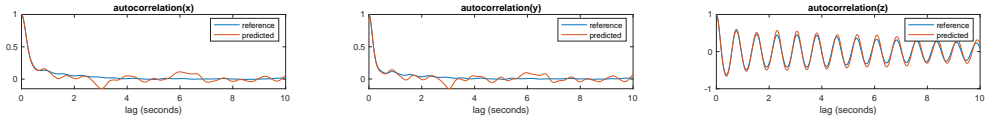}
		\caption{Ex. 1: Lorenz 63 Full System -- Autocorrelation function comparison.}
		\label{fig:lorenz63full_autocorr}
	\end{center}
\end{figure}

Prediction is carried out to $T=100$ seconds ($10,000$ time steps)
from a new initial condition $(10,10,20)$. Figure
\ref{fig:lorenz63full_error} shows the trajectory prediction as well
as the log absolute error. We see that despite how the prediction
trajectories quickly deviate from the reference trajectories (as
should be expected for a chaotic system), the
pointwise error in all three variables remains bounded over the long
term. In addition, Figures \ref{fig:lorenz63full_phase},
\ref{fig:lorenz63full_hist}, and \ref{fig:lorenz63full_autocorr}, show
qualitatively similar phase plots, histograms, and autocorrelation
functions, with at worst the quantitative chaos statistics of 
 $9\%$ in relative error. See Table \ref{table:63-full} for details.
\begin{table} [htbp]
\begin{center}
\begin{tabular}{| c | c | c | c |}
\hline
 Metrics & Reference solution & 
   DNN prediction & Relative error \\
\hline
  Correlation dimension & 1.9924 & 2.1722 & 9.0\% \\
  \hline
 Approximate entropy & 0.0435 & 0.0441 & 1.4\% \\
\hline
  Lyapunov exponent & 0.5504  & 0.5534 & 0.5\%\\
\hline
\end{tabular}
\caption{Ex. 1: Lorenz 63 Full System -- Metrics for chaotic behavior comparison.}
\label{table:63-full}
\end{center}
\end{table}

\subsubsection{Example 2: Reduced 1- or 2-dimensional systems}

In this example we consider the six possible combinations of
partially-observed systems arising from Lorenz 63. Specifically, we
consider observing two variables of only $x$ and $y$, only $x$ and $z$, and only $y$
and $z$, as well as one variable of only $x$, only $y$, and only $z$. When variables
are missing, the Mori-Zwanzig formulation informs us that memory is
required to learn the reduced system dynamics. Hence, in all of the
following experiments a trajectory of the full Lorenz system is
generated by a high-order numerical solver and the observed variables
are observed at every $\Delta t=0.01$ starting from initial condition
$(1,1,1)$ until $T=10,000$ seconds. (The unobserved variables are
discarded from the true system solutions.) From this long trajectory,
$10,000$ data sequences of length $0.2$ seconds ($21$ time steps) are
taken from only the observed variable(s) as training data. This allows
$0.1$ seconds ($10$ steps) for memory and $0.1$ seconds ($10$ steps)
for recurrent loss. A standard ResNet with $3$ hidden layers and $20$
neurons each is used, and the mean squared recurrent loss function is
minimized using Adam with a constant learning rate of $10^{-3}$ for
$10,000$ epochs. 

\begin{figure}[htbp]
	\begin{center}
		\includegraphics[width=\textwidth]{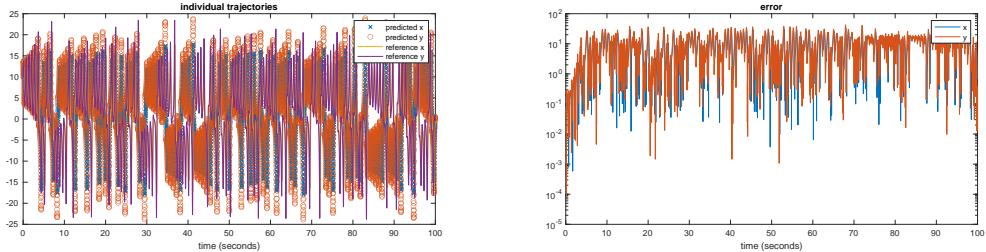}
		\caption{Ex. 2: Lorenz 63 Reduced System of $x$ and $y$ -- Individual trajectory comparison and pointwise error.}
		\label{fig:lorenz63reduced12_error}
	\end{center}
\end{figure}

\begin{figure}[htbp]
	\begin{center}
		\includegraphics[width=\textwidth]{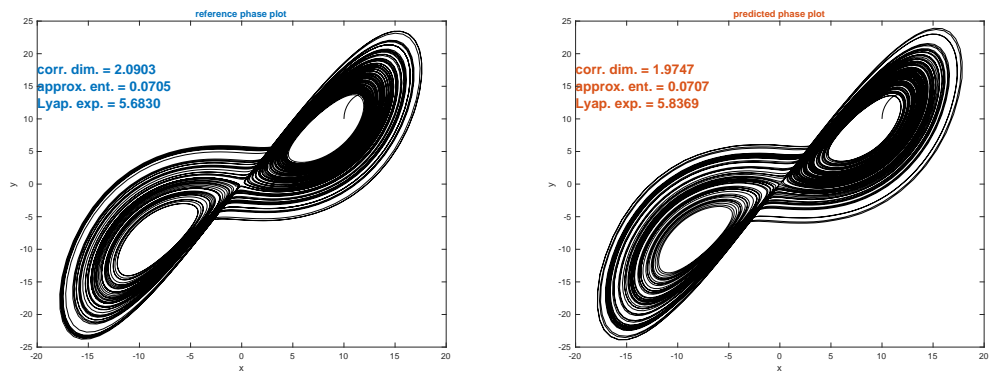}
		\caption{Ex. 2: Lorenz 63 Reduced System of $x$ and $y$ -- Phase
                  plots. (Left: reference; Right: DNN prediction.)}
		\label{fig:lorenz63reduced12_phase}
	\end{center}
\end{figure}

\begin{figure}[htbp]
	\begin{center}
		\includegraphics[width=\textwidth]{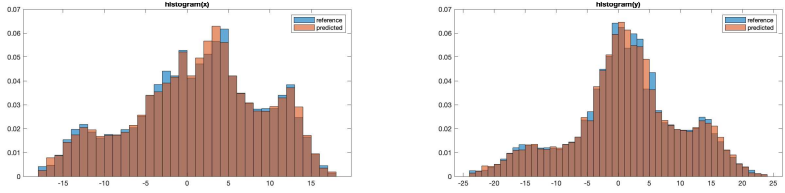}
		\caption{Ex. 2: Lorenz 63 Reduced System of $x$ and $y$ -- Histogram comparison.}
		\label{fig:lorenz63reduced12_hist}
	\end{center}
\end{figure}

\begin{figure}[htbp]
	\begin{center}
		\includegraphics[width=\textwidth]{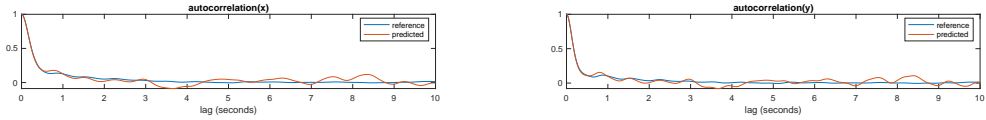}
		\caption{Ex. 2: Lorenz 63 Reduced System of $x$ and $y$ -- Autocorrelation function comparison.}
		\label{fig:lorenz63reduced12_autocorr}
	\end{center}
\end{figure}

Prediction for the observed variables is carried out to $T=100$
seconds ($10,000$ time steps) from a new initial condition
$(10,10,20)$, with the corresponding observables in different
cases. The two-variable $x$ and $y$ system is shown in Figures
\ref{fig:lorenz63reduced12_error}, \ref{fig:lorenz63reduced12_phase},
\ref{fig:lorenz63reduced12_hist}, and
\ref{fig:lorenz63reduced12_autocorr}. We see bounded pointwise error
indicating stability, qualitatively similar phase plots indicating
similar behavior, similar histogram indicating the appropriate
density, and chaos statistics very close to the reference.

\begin{figure}[htbp]
	\begin{center}
		\includegraphics[width=\textwidth]{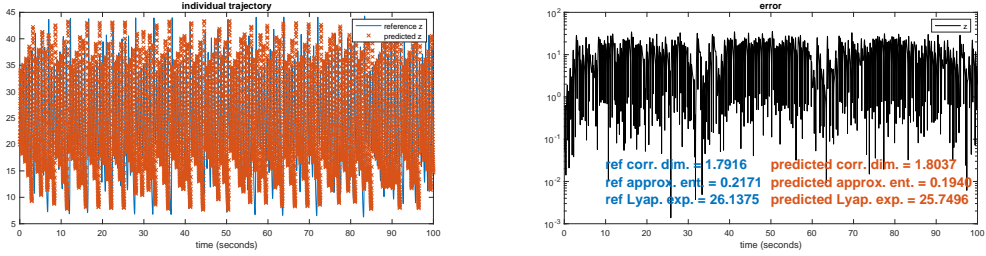}
		\caption{Ex. 2: Lorenz 63 Reduced System of $z$ -- Individual trajectory and chaos statistics comparison, and pointwise error.}
		\label{fig:lorenz63reduced3_error}
	\end{center}
\end{figure}

\begin{figure}[htbp]
	\begin{center}
		\includegraphics[width=\textwidth]{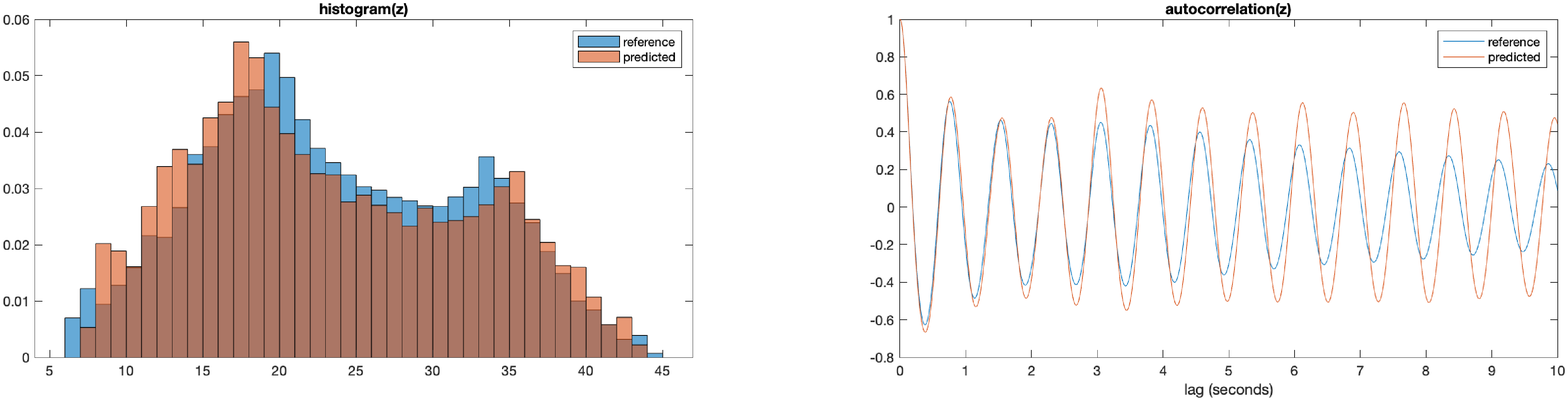}
		\caption{Ex. 2: Lorenz 63 Reduced System of $z$ -- Histogram and autocorrelation function comparison.}
		\label{fig:lorenz63reduced3_hist}
	\end{center}
\end{figure}

The one-variable $z$ system is shown in Figures
\ref{fig:lorenz63reduced3_error} and
\ref{fig:lorenz63reduced3_hist}. Once again we see bounded pointwise
error, similar histograms and autocorrelation functions, as well as
very accurate chaos statistics. The correlation dimension, approximate
entropy, and Lyapunov exponent values for all of the combinations
of observed variables are reported in Table \ref{table:63-reduced}. The
predicted statistics are at worst $17.5\%$ off from the
reference statistics in relative error, but are typically significantly lower, and we note that in the few cases where relative error exceeds $10\%$ it is only in one of the three statistics with the other two being much more accurate. E.g., we see that the reduced system of $x$ and $z$ has a relative error in Lyapunov exponent of $17.5\%$, but matches correlation dimension and approximate entropy at relative errors of $0.1\%$ and $3.0\%$.

\begin{table} [htbp]
\begin{center}
\begin{tabular}{| c | c | c | c |}
\hline
 Metrics & Reference solution ($x$,$y$) & DNN prediction ($x$,$y$) & Relative error \\
\hline
  Correlation dimension & 2.0903 & 1.9747 & 5.5\% \\
  \hline
 Approximate entropy & 0.0705 & 0.0707 & 0.3\% \\
\hline
  Lyapunov exponent & 5.6830  & 5.8369 & 2.7\%\\
\hline
\hline
 Metrics & Reference solution ($x$,$z$) & DNN prediction ($x$,$z$) & Relative error \\
\hline
  Correlation dimension & 2.1484 & 2.1472 & 0.1\% \\
  \hline
 Approximate entropy & 0.0796 & 0.0820 & 3.0\% \\
\hline
  Lyapunov exponent & 4.6859  & 5.5066 & 17.5\%\\
\hline
\hline
Metrics & Reference solution ($y$,$z$) & DNN prediction ($y$,$z$) & Relative error \\
\hline
  Correlation dimension & 2.2618 & 2.1986 & 2.8\% \\
  \hline
 Approximate entropy & 0.0564 & 0.0571 & 1.2\% \\
\hline
  Lyapunov exponent & 3.1263  & 3.5128 & 12.4\%\\
\hline
\hline
Metrics & Reference solution ($x$) & DNN prediction ($x$) & Relative error \\
\hline
  Correlation dimension & 2.0160 & 1.8821 & 6.6\% \\
  \hline
 Approximate entropy & 0.1956 & 0.1986 & 1.5\% \\
\hline
  Lyapunov exponent & 27.0396  & 27.3691 & 1.2\%\\
\hline
\hline
Metrics & Reference solution ($y$) & DNN prediction ($y$) & Relative error \\
\hline
  Correlation dimension & 1.7573 & 1.7979 & 2.3\% \\
  \hline
 Approximate entropy & 0.2053 & 0.2027 & 1.3\% \\
\hline
  Lyapunov exponent & 29.1866  & 29.5374 & 1.2\%\\
\hline
\hline
Metrics & Reference solution ($z$) & DNN prediction ($z$) & Relative error \\
\hline
  Correlation dimension & 1.7916 & 1.8037 & 0.7\% \\
  \hline
 Approximate entropy & 0.2171 & 0.1940 & 10.6\% \\
\hline
  Lyapunov exponent & 26.1375  & 25.7496 & 1.5\%\\
\hline
\end{tabular}
\caption{Ex. 2: Lorenz 63 Reduced Systems -- Metrics for chaotic behavior comparison.}
\label{table:63-reduced}
\end{center}
\end{table}

\subsection{High-dimensional System: Lorenz 96}

The Lorenz 96 system \cite{lorenz1996predictability} is an $N$-dimensional dynamical system given by
\begin{align}
\frac{dx_i}{dt} &= (x_{i+1}-x_{i-2})x_{i-1} + F
\end{align}
for $i=1,\ldots,N$, for $N\ge 4$ where $x_{-1}=x_{N-1}$, $x_0=x_N$, and $x_{N+1}=x_1$ and $F$ is a forcing parameter. If $F=8$, the system exhibits chaotic behavior. In the examples below, we choose $N=40$ as in \cite{lorenz1998optimal}.

\subsubsection{Example 3: Full 40-dimensional system}

In this example all $40$ state variables are observed. In particular, a trajectory is generated by a high-order numerical solver with $\Delta t=0.01$ from initial condition $(8.0081,8,8,\ldots,8)$ until $T=10,000$ seconds ($1,000,000$ time steps). From this long trajectory, $100,000$ chunks of length $0.1$ seconds ($11$ time steps) are taken as training data. This allows $10$ steps to be used for recurrent loss. A standard ResNet with $3$ hidden layers with $200$ neurons each is used, and the mean squared recurrent loss function is minimized using Adam with a constant learning rate of $10^{-3}$ for $2,000$ epochs.

\begin{figure}[htbp]
	\begin{center}
		\includegraphics[width=\textwidth]{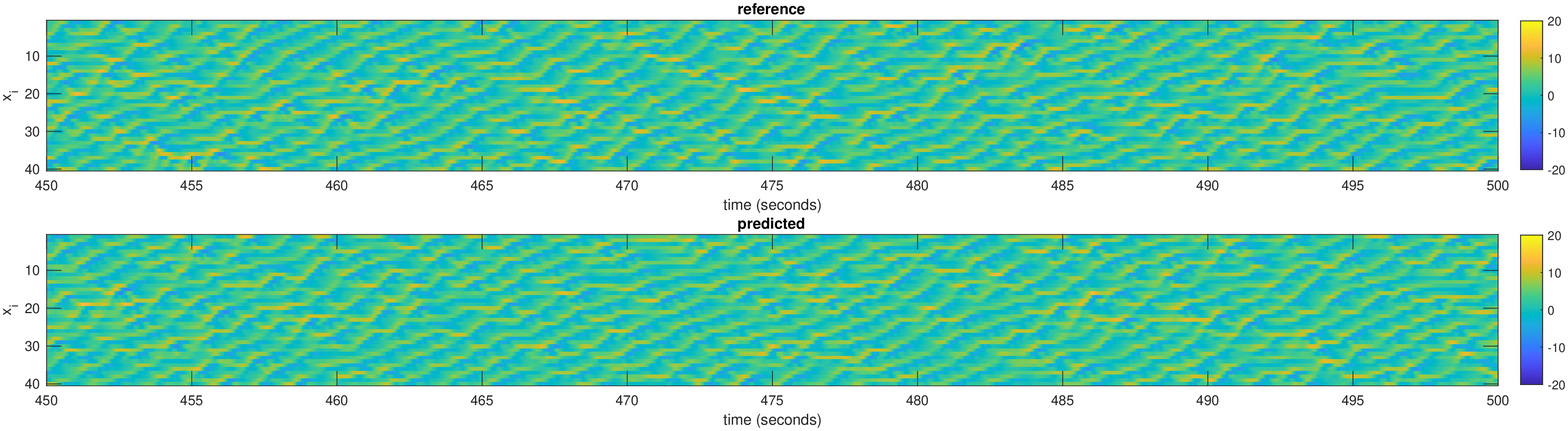}
		\caption{Ex. 3: Lorenz 96 Full System -- Individual trajectory comparison.}
		\label{fig:lorenz96full}
	\end{center}
\end{figure}

Prediction is carried out to $T=500$ ($50,000$ time steps) from new initial condition $(8.01,8,8,\ldots,8)$. Figure \ref{fig:lorenz96full} shows the prediction results for $t=450$ through $t=500$ with the $i$th  row corresponding to the trajectory of the variable $x_i$. We see the same type of qualitative behavior in the reference and prediction, with wave-like forms flowing through the $40$ variables. Table \ref{table:2} show the correlation dimension, approximate entropy, and Lyapunov exponent for this experiment. As in Ex. 2, we note that while the relative error for approximate entropy is very high at $46.3\%$, we consider that it may be an anomaly as the other two statistics are significantly more accurate at $3.1\%$ and $3.5\%$.

\begin{table} [htbp]
\begin{center}
\begin{tabular}{| c | c | c | c |}
\hline
 Metrics & Reference solution & 
   DNN prediction & Relative error \\
\hline
  Correlation dimension & 1.4025 & 1.3585 & 3.1\% \\
  \hline
 Approximate entropy & 0.0067 & 0.0036 & 46.3\% \\
\hline
  Lyapunov exponent & 0.2419  & 0.2506 & 3.6\%\\
\hline
\end{tabular}
\caption{Ex. 3: Lorenz 96 Full System -- Metrics for chaotic behavior comparison.}
\label{table:2}
\end{center}
\end{table}

\subsubsection{Example 4: Reduced 3-dimensional system}

In this example we consider observing only $x_1$, $x_2$, and $x_3$ from the 40-dimensional Lorenz 96 system. Again, Mori-Zwanzig informs us that memory is required to learn the reduced system dynamics. Hence, a trajectory of the full system is generated by a high-order numerical solver with $\Delta t=0.01$ from initial condition $(8.0081,8,8,\ldots,8)$ until $T=10,000$ seconds. From this long trajectory, $10,000$ chunks of length $1.1$ seconds ($111$ time steps) are taken from only the observed variables as training data. This allows $1$ second ($100$ time steps) for memory and $0.1$ seconds ($10$ time steps) for recurrent loss. A standard ResNet with $10$ hidden layers and $20$ neurons each is used, and the mean squared recurrent loss function is minimized using Adam with a constant learning rate of $10^{-3}$ for $10,000$ epochs.

\begin{figure}[htbp]
	\begin{center}
		\includegraphics[width=\textwidth]{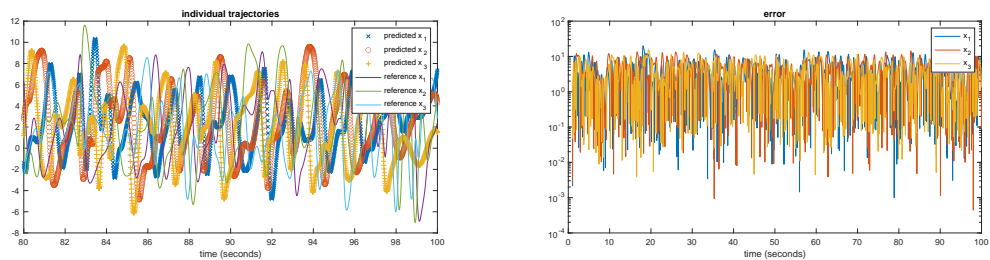}
		\caption{Ex. 4: Lorenz 96 Reduced System of $x_1$, $x_2$, $x_3$ -- Individual trajectory comparison and pointwise error.}
		\label{fig:lorenz96reduced123_error}
	\end{center}
\end{figure}

\begin{figure}[htbp]
	\begin{center}
		\includegraphics[width=\textwidth]{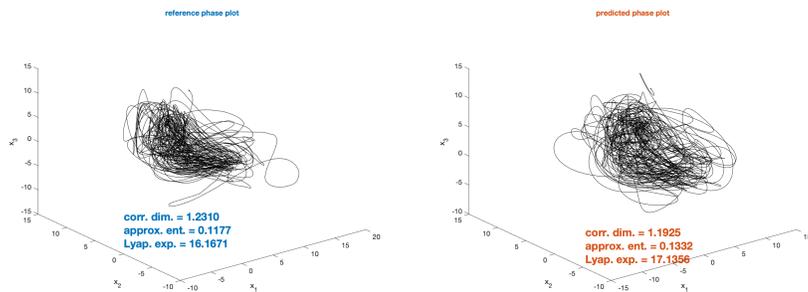}
		\caption{Ex. 4: Lorenz 96 Reduced System of $x_1$, $x_2$, $x_3$ -- Phase
                  plots. (Left: reference; Right: DNN prediction.)}
		\label{fig:lorenz96reduced123_phase}
	\end{center}
\end{figure}

\begin{figure}[htbp]
	\begin{center}
		\includegraphics[width=\textwidth]{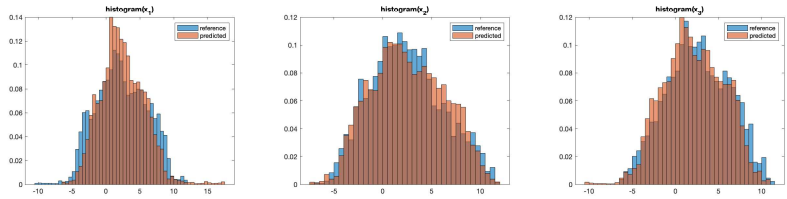}
		\caption{Ex. 4: Lorenz 96 Reduced System of $x_1$, $x_2$, $x_3$ -- Histogram comparison.}
		\label{fig:lorenz96reduced123_hist}
	\end{center}
\end{figure}

\begin{figure}[htbp]
	\begin{center}
		\includegraphics[width=\textwidth]{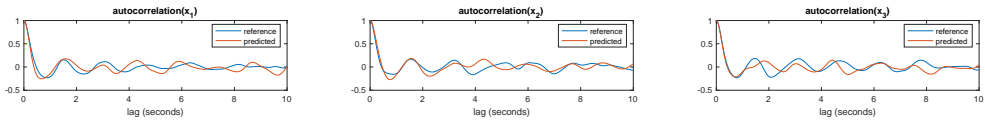}
		\caption{Ex. 4: Lorenz 96 Reduced System of $x_1$, $x_2$, $x_3$ -- Autocorrelation function comparison.}
		\label{fig:lorenz96reduced123_autocorr}
	\end{center}
\end{figure}

Prediction is carried out to $T=100$ ($10,000$ time steps) from new initial condition $(8.01,8,8,\ldots,8)$. Figure \ref{fig:lorenz96reduced123_error} shows the individual trajectories between $t=80$ and $t=100$ seconds along with bounded pointwise error, which again indicates stability of the prediction. Figure \ref{fig:lorenz96reduced123_phase} shows the phase plots and quantitative measures, while Figures \ref{fig:lorenz96reduced123_hist} and \ref{fig:lorenz96reduced123_autocorr} show the histogram and autocorrelation function comparisons. Due to the significantly more chaotic and complex nature of the Lorenz 96 system, it is now more difficult to pick out identifying characteristics in the trajectories and phase plots to compare. Nevertheless, the chaos statistics reported in Table \ref{table:96-reduced} indicate a strong match with only one of three statistics exceeding $10\%$ relative error.

\begin{table} [htbp]
\begin{center}
\begin{tabular}{| c | c | c | c |}
\hline
 Metrics & Reference solution & 
   DNN prediction & Relative error \\
\hline
  Correlation dimension & 1.2310 & 1.1925 & 3.1\% \\
  \hline
 Approximate entropy & 0.1177 & 0.1332 & 13.1\% \\
\hline
  Lyapunov exponent & 16.1671  & 17.1356 & 6.0\%\\
\hline
\end{tabular}
\caption{Ex. 4: Lorenz 96 Reduced System -- Metrics for chaotic behavior comparison.}
\label{table:96-reduced}
\end{center}
\end{table}

%% file: Conclusion.tex
\section{Conclusion} \label{sec:conclusions}

We have presented a systematic and rigorous examination of learning the flow maps of fully- and partially-observed chaotic systems using an approachable DNN framework. While in most papers that examine this topic, pointwise error or a visual comparison of phase plots are used to assess the accuracy of network prediction, here we use these tools as well as a variety of other measures to demonstrate accurate long-term chaotic behavior prediction. Our numerical examples show that this DNN framework is able to learn long-term chaotic behavior even when systems are severely under-observed and training data are collected from a single initial condition.